\documentclass[11pt]{article}

\usepackage[final]{acl}

\usepackage{times}
\usepackage{latexsym}

\usepackage[T1]{fontenc}

\usepackage[utf8]{inputenc}

\usepackage{microtype}
\usepackage{inconsolata}
\usepackage{graphicx}

\usepackage{tabularx}
\usepackage{array}
\usepackage{xcolor}
\usepackage{multirow}
\usepackage{tikz}
\usetikzlibrary{shapes, positioning, arrows.meta, calc}
\usepackage{fontawesome5}
\usepackage{float}
\usepackage{booktabs}
\usepackage{afterpage}

\title{How Benchmarks Mis-Score Computer-Use Agents}

\author{
  Zihan Dong \\
  Georgia Institute of Technology \\
  \texttt{zdong312@gatech.edu}
  \And
  Zhiyuan Ma \\
  North Carolina State University \\
  \texttt{zma24@ncsu.edu}
  \And
  Zekun Wang \\
  Georgia Institute of Technology \\
  \texttt{zekun@gatech.edu}
  \And
  Yunqing Li \\
  Lenovo AI Technology Center (LATC) \\
  \texttt{yli59@lenovo.com}
  \And
  Zirou Liu \\
  UNC at Chapel Hill \\
  \texttt{zirliuyy@unc.edu}
  \And
  Ruixuan Deng \\
  Georgia Institute of Technology \\
  \texttt{rdeng62@gatech.edu}
  \And
  Qishi Zhan \\
  Marquette University 
  \texttt{qishizhan7@gmail.com}
  \And
  Rui Qian \\
  Fudan University \\
  \texttt{qiianruii@gmail.com} \\
}

\author{
\textbf{Zihan Dong}$^{1}$,
\textnormal{Zhiyuan Ma}$^{2}$,
\textnormal{Zekun Wang}$^{1}$,
\textnormal{Yunqing Li}$^{3}$, \\
\textnormal{Zirou Liu}$^{4}$,
\textnormal{Ruixuan Deng}$^{1}$,
\textnormal{Qishi Zhan}$^{5}$,
\textnormal{Rui Qian}$^{6}$ \\[4pt]
\normalfont
$^{1}$Georgia Institute of Technology \\
$^{2}$North Carolina State University \\
$^{3}$Lenovo AI Technology Center \\
$^{4}$University of North Carolina at Chapel Hill \\
$^{5}$Marquette University \\
$^{6}$Fudan University \\[4pt]
\texttt{zdong312@gatech.edu}
}

\begin{document}
\maketitle

\begin{abstract}
Computer-use agents (CUA) are being deployed to browse the web and operate desktop software, yet their benchmark scores are still commonly produced by brittle scripted oracles. A score is the output of a pipeline in which tasks can be stale, trajectories can omit decisive visual evidence, evaluators can reject valid alternatives, and aggregate reports can hide the cause of failure. We organize these problems into a reliability framework spanning task construction, trajectory observation, scoring, and reporting. We then audit 150 public failure-scored trajectories from five web, enterprise-workflow, and desktop-control benchmarks, find that 15.3\% of FAIL verdicts are wrong: 10.7\% are evaluator false negatives and 4.7\% are broken tasks. For genuine failures, a three-tier diagnostic taxonomy shows that verification/feedback and planning failures dominate execution/grounding errors, while a single scalar success rate can not explain. We connect these findings to newer long-horizon CUA benchmarks and derive stage-specific design rules for CUA evaluation.
\end{abstract}

\section{Introduction}

Computer-use agents (CUA) are already being placed between users and consequential web and desktop interfaces: they search, fill forms, manipulate files, and complete multi-application workflows~\cite{openaiOperator2025,googleAiModeAgentic2025,qwenAgentic2026,doubaoPhoneAssistant2025}. Their evaluation, however, still depends heavily on scripted oracles that inspect a URL, a DOM field, or a final machine state. The deployment--measurement gap becomes sharper as tasks grow from the roughly 30-action episodes of OSWorld to hour-scale, cross-application workflows: OSWorld~2.0 reports a median human completion time of about 1.6 hours, while Odysseys and WeaveBench explicitly target long-horizon, multi-site or hybrid-interface work~\cite{osworld2024,osworld2_2026,odysseys2026,weavebench2026}.

These settings cannot be measured faithfully by treating execution as self-validating. A useful solution may follow an unforeseen path; a final state may look correct while violating an intermediate constraint; and an agent may fail because the live site, virtual machine, or checker is wrong rather than because its CUA capability is weak. Newer benchmarks have therefore begun to add graded rubrics, trajectory-aware judges, and separate partial-completion signals~\cite{odysseys2026,weavebench2026,cuaRewardBench2026}. These developments make evaluator reliability more current, not less: each additional judge is another measurement instrument that must be validated.

\begin{figure*}[t]
\centering
\includegraphics[width=\textwidth]{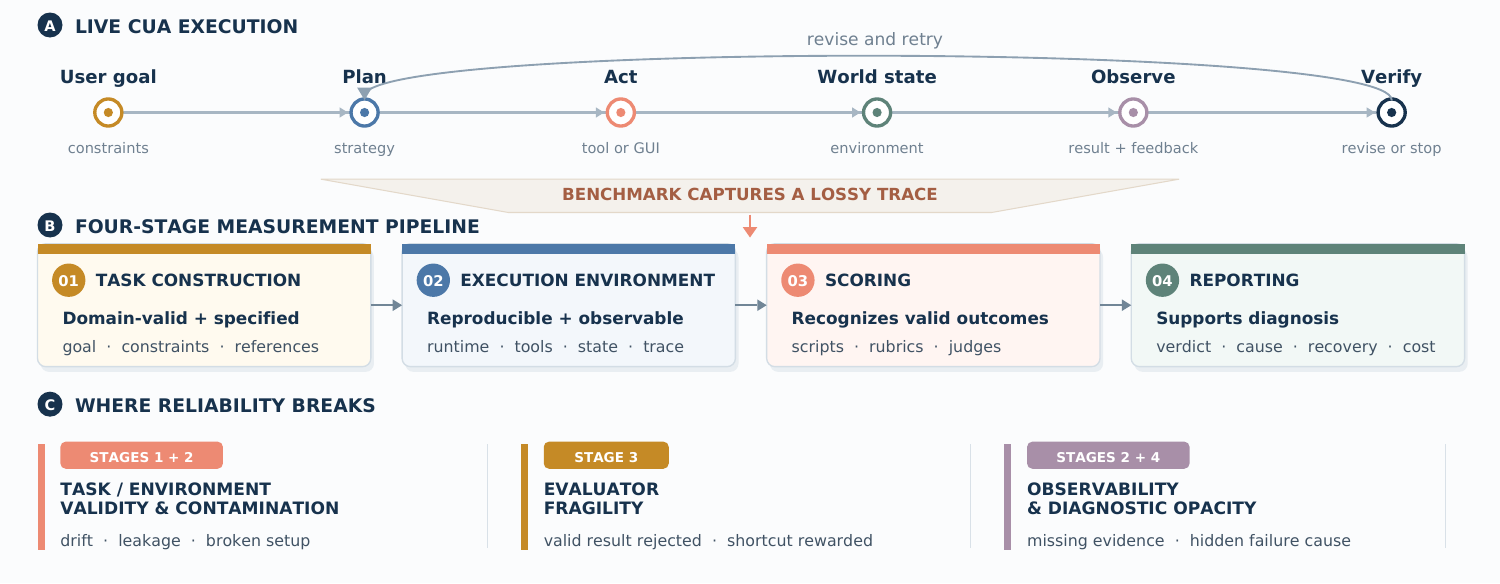}
\caption{\textbf{A CUA benchmark score is a pipeline output, not a direct observation of capability.}
The benchmark projects the execution loop into four measurement stages. Task validity and contamination
enter during construction, evaluator fragility enters during scoring, and observability and diagnostic
opacity span trajectory observation and reporting.}
\label{fig:audit-framework}
\end{figure*}

The central problem is that an agent score is the output of a \textit{pipeline}, not a direct observation of capability. Stale or leaked tasks, broken environments, incomplete trajectories, scoring scripts that reject valid alternatives, and aggregate success rates that obscure failure causes all distort evaluation. These are not interchangeable ``agent failures.'' They arise at distinct stages of task construction, observation, scoring, and reporting. Collapsing them into a single category produces leaderboards that mis-rank systems and directs engineering effort toward the wrong component~\cite{webarenaVerifiedGithub,agentRewardBench2025,searchTimeContam2025}. We therefore study the \textbf{reliability of the CUA evaluation pipeline}. Figure~\ref{fig:audit-framework} provides the paper's organizing schema. We first locate three threats---contamination and task invalidity, evaluator fragility, and diagnostic opacity---at the stages where they enter (\S\ref{sec:systemic-failures}). We then examine their current manifestations across web, enterprise-workflow, and desktop evaluation (\S\ref{sec:domain-audit}), and test them directly by re-auditing 150 public CUA trajectories (\S\ref{sec:taxonomy}). We expect readers can trace a CUA verdict through the full evaluation pipeline, distinguish evaluator false negatives and broken tasks from genuine agent failures, and diagnose those failures with a three-tier codebook adapted from MAST~\cite{cemri2025multi}.

Our audit of 150 failure-scored trajectories across five benchmarks shows why these distinctions matter: 15.3\% of FAIL verdicts are wrong (10.7\% evaluator false negatives and 4.7\% broken tasks), while genuine failures are dominated by feedback-blindness and planning errors. We translate these findings into stage-specific guidance for task construction, environments, scoring, and reporting. More broadly, we hope this reliability perspective inspires further CUA innovation, evaluation reform, and evidence-based governance and regulation.
Our scope focused on the evaluation of CUA across current web, desktop/OS, enterprise-workflow, and mobile tasks. Prior surveys organize evaluation by capability, application, or methodology~\cite{yehudai2025survey,surveyAgents2025}; they ask \textit{what evaluation resources exist}, whereas this paper asks \textit{when a CUA verdict should be distrusted} and supports that diagnosis with trajectory-level evidence.

\section{Evaluation Reliability Framework}
\label{sec:systemic-failures}

We framed CUA evaluation as a four-stage measurement pipeline: task construction, trajectory observation, scoring, and reporting. Reliability depends on solvable tasks that are independent of system development, evidence-complete trajectories, evaluators that recognize acceptable outcomes, and reports that support diagnosis. Figure~\ref{fig:audit-framework} locates three recurring threats across these stages.

\subsection{Task Validity and Contamination}

A CUA benchmark produces credible rankings when its tasks, environments, and hidden references remain valid and independent of system development. Live sites drift, credentials expire, VM images lack assumed hardware, and professional software changes its menus or file formats. 
Search-Time Contamination (STC) could happen when public instructions and reference artifacts enter training or benchmark-specific tuning, while web-enabled agents can retrieve evaluation-specific instructions, answers, or near-duplicate artifacts during a run, so a score can reflect access to benchmark material rather than the intended interaction capability~\cite{searchTimeContam2025}. Living task pools such as Agents' Last Exam and state-rich suites such as OSWorld~2.0 reduce staleness and saturation pressure; explicit versioning of tasks, environments, and graders makes these changes traceable~\cite{agentsLastExam2026,osworld2_2026}.

Mitigation follows the CUA threat model. Closed self-hosted sites can disallow external retrieval and whitelist network access. On the live web, lookup is part of the capability, so time-stamped tasks and ground truth track freshness directly. In closed enterprise settings, the risks shift from external STC to \textit{internal leakage} (answers in the retrieval index) and \textit{process leakage} (evaluation logs reused in training)~\cite{searchTimeContam2025}.

\subsection{Evaluator Fragility}

Open-ended CUA tasks admit multiple valid outcomes and paths. Hard-coded checks of a string, URL, DOM node, or machine state can reject acceptable alternatives; outcome-only checks can instead reward a superficially correct artifact produced through invalid shortcuts~\cite{webarenaVerifiedGithub,weavebench2026}. ``Verified'' subsets improve task and evaluator quality while exposing noise in unaudited automation. Semantic, rubric-based, and judge-based matching broaden coverage but add measurement instruments whose reliability depends on independent calibration~\cite{agentAsJudge2024,agentRewardBench2025,cuaRewardBench2026}.

\subsection{Observability and Diagnostic Opacity}

A scalar success rate cannot distinguish a wrong plan from a wrong click, a stale environment, or a checker error. Logs that omit screenshots, tool responses, timestamps, or state transitions can make post-hoc diagnosis impossible, creating an observability failure before taxonomy is applied. Replayable, screenshot-inclusive trajectories preserve the evidence, and a shared vocabulary localizes failure within the CUA loop~\cite{surveyAgents2025,cemri2025multi,osworld2024}. \S\ref{sec:taxonomy} operationalizes this sequence by auditing verdict validity before classifying genuine failures by stage.

\section{CUA Evaluation Landscape}
\label{sec:domain-audit}

We organize the discussion around the three task settings summarized in
Table~\ref{tab:cua-evaluation-landscape}, focusing on the cross-cutting
demands that distinguish them.

\begin{table*}[t]
\centering
\scriptsize
\setlength{\tabcolsep}{3pt}
\renewcommand{\arraystretch}{1.05}
\renewcommand{\tabularxcolumn}[1]{m{#1}}
\begin{tabularx}{\textwidth}{@{}
  >{\hsize=.40\hsize\linewidth=\hsize\centering\arraybackslash}X|
  >{\hsize=.58\hsize\linewidth=\hsize\centering\arraybackslash}X|
  >{\hsize=1.37\hsize\linewidth=\hsize\arraybackslash}X|
  >{\hsize=1.65\hsize\linewidth=\hsize\arraybackslash}X@{}}
\hline
\textbf{Task setting} & \textbf{Application} & \textbf{Metric / evaluation focus} & \textbf{Representative benchmarks} \\ \hline
\multirow[c]{4}[0]{=}[-3.0\baselineskip]{\centering\textbf{Specialized professional GUI}} & Healthcare administration and clinical software & \textbf{Workflow validity and safety:} Deterministic task/subtask completion, correct cross-system state, and clinically consequential error checks. & HealthAdminBench~\cite{healthadminbench2026}, MedCUA-Bench~\cite{medcuabench2026}, ProSoftArena~\cite{prosoftarena2025} \\ \cline{2-4}

 & CAD and electronic design automation & \textbf{Professional GUI execution:} Screenshot-to-action accuracy and successful operation of dense, high-resolution industrial CAD interfaces. & GUI-EDA~\cite{guieda2025}, CADWorld~\cite{dong2026cadworld}, ProSoftArena~\cite{prosoftarena2025}, GUI-vs-CLI~\cite{guivscli2026}, ScreenSpot-Pro~\cite{screenspotpro2025} \\ \cline{2-4}

 & Scientific instruments and data workflows & \textbf{State- and result-based completion:} Subtask and end-to-end success for instrument control, scientific software, and data-pipeline workflows. & LabOSBench~\cite{labosbench2026}, ScienceBoard~\cite{scienceboard2026}, Spider2-V~\cite{spider2v2024}, ProSoftArena~\cite{prosoftarena2025} \\ \cline{2-4}

 & Creative production & \textbf{Artifact and interaction quality:} Completion of long-horizon image, design, video, audio, and 3D workflows, including clarification and revision. & DeskCraft~\cite{deskcraft2026}, PSBench~\cite{psbench2026}, ProSoftArena~\cite{prosoftarena2025}, ScreenSpot-Pro~\cite{screenspotpro2025} \\ \hline

\multirow[c]{2}[0]{=}[-1.0\baselineskip]{\centering\textbf{Web navigation}} & Dynamic navigation & \textbf{Semantic success:} Type-aware exact matching with semantic normalization (handling DOM mutations). & WebArena \cite{webarena2023,webarenaVerifiedGithub}, Mind2Web \cite{mind2web2023}, MiniWoB++ \cite{miniwobplusplus2018}, AssistantBench~\cite{agentRewardBench2025}, Odysseys~\cite{odysseys2026} \\ \cline{2-4}

 & Enterprise workflows & \textbf{Visual and workflow completion:} Visual understanding of elements; completion of administrative forms. & VisualWebArena \cite{visualwebarenaRepo,visualwebarena2024}, WebVoyager \cite{webvoyager2024}, WebLINX \cite{weblinx2024}, WorkArena \cite{workarena2024,workarenaRepo} \\ \hline

\multirow[c]{2}[0]{=}[-0.55\baselineskip]{\centering\textbf{Device control}} & Desktop control & \textbf{Execution-based snapshot:} Final file-system state verification across Ubuntu/Windows apps. & OSWorld \cite{osworld2024}, OSWorld~2.0 \cite{osworld2_2026}, Agents' Last Exam \cite{agentsLastExam2026}, WeaveBench~\cite{weavebench2026} \\ \cline{2-4}

 & Mobile interaction & \textbf{Cross-app workflow:} Success rate on multi-app tasks; pixel-coordinate precision (GUI grounding). & MobileWorld \cite{mobileworld2025}, AndroidWorld \cite{androidworld2024}, Mobile-Env~\cite{mobileenv2023} \\ \hline

\end{tabularx}
\caption{CUA benchmark landscape by task setting, application, evaluation focus, and representative benchmarks.}
\label{tab:cua-evaluation-landscape}
\end{table*}

\subsection{Specialized Professional GUI}

Professional interfaces make domain state part of the task. A visually
plausible action can violate a workflow dependency, change the wrong
structured object, or produce an artifact that cannot support later
work. The generalization problem is thus not
transfer to an unfamiliar layout but transfer across domain
conventions, latent dependencies, and representations with no visible
equivalent on screen
~\cite{medcuabench2026,dong2026cadworld,labosbench2026,deskcraft2026}.

This setting also separates interface fluency from professional
competence. Static target localization can establish whether an agent
can operate a dense interface, but end-to-end use additionally requires
maintaining domain intent while the artifact and application state evolve
over many actions~\cite{screenspotpro2025,prosoftarena2025}. Thus, progress
in generic GUI grounding does not by itself imply reliable transfer to
professional work.

\subsection{Web Navigation}

Web tasks are distinguished by remote, mutable state rather than by the
browser alone. The same action can have different consequences depending
on authentication, session history, asynchronous updates, and changes
made by other services. Moreover, many goals admit several legitimate
routes, so success depends on preserving intent across navigation choices
instead of reproducing a single interaction sequence
~\cite{webarena2023,browsergym2024,workarena2024}.

The interface representation is likewise incomplete from any single
view. DOM structure exposes semantics that pixels may obscure, whereas
screenshots retain visual relationships absent from accessibility trees.
Agents must reconcile these views while pages reflow and content changes,
making web navigation a problem of state interpretation as much as
element selection~\cite{visualwebarena2024,weblinx2024}. Longer,
multi-site tasks amplify this coupling because early retrieval and
navigation choices become inputs to much later actions
~\cite{odysseys2026}.

\subsection{OS and Mobile}

OS-level tasks compose applications through persistent machine state.
Information may move from a page to a clipboard, file, application
object, or system setting, and later steps may depend on transformations
that are no longer visible. Cross-application competence therefore
requires continuity of state and constraints across tool boundaries, not
just competence within each interface~\cite{osworld2024,osworld2_2026,
weavebench2026}.

Desktop and mobile control share this persistence but impose different
interaction bottlenecks. Desktop workflows expose more simultaneous
state and richer tool combinations; mobile workflows constrain the
viewport and route more transitions through OS-managed surfaces.
In both cases, small local mistakes can propagate because subsequent
applications consume the state produced earlier, turning recovery and
state reconstruction into central long-horizon capabilities
~\cite{androidworld2024,mobileworld2025}.

\section{Empirical Audit of Benchmark Verdicts}
\label{sec:taxonomy}

To determine when a benchmark verdict should be distrusted and what a
binary failure score conceals, we audit 150 failure-scored trajectories
from five CUA benchmarks. For each trajectory, we first assess whether
the recorded failure reflects an agent failure, an evaluator false
negative, a broken task, or insufficient evidence; only genuine agent
failures are then assigned a process-level diagnosis. Two vision-enabled
LLM judges independently review the complete trajectories, including
reasoning, actions, and screenshots. We anchor their judgments with two
human groups that label the same review set independently and blind to
the LLM outputs and to each other. The remainder of this section presents
the audit setting, labeling procedure, diagnostic taxonomy, and results
in that order.

\subsection{Audit Setting}

We sampled public trajectories with step-level reasoning, actions, and
screenshots using a deterministic, stratified procedure (random seed
20260717). From AgentRewardBench~\cite{agentRewardBench2025}, we formed
unique (benchmark, task, agent, experiment) tuples, shuffled within each
benchmark--agent stratum, round-robin interleaved agent strata, and kept
only trajectories with released cumulative reward zero until reaching
quotas of 25 WebArena, 25 WorkArena, 24 VisualWebArena, and 24
AssistantBench trajectories. From OSWorld-Verified, we took 20
zero-reward trajectories from each of three released agent runs,
shuffling within application domains and round-robin interleaving the
domains. This produced 158 trajectories; eight (one per OSWorld run and
five spread across the web benchmarks and agent models) were selected
for codebook calibration and excluded from analysis. The remaining 150
comprise 57 OSWorld trajectories and 93 AgentRewardBench trajectories:
23 each from WebArena, VisualWebArena, and AssistantBench, and 24 from
WorkArena. We perform no new agent runs. Because every sampled unit was
released as FAIL, the audit estimates false negatives among recorded
failures but not false positives among successes.

The five benchmarks also expose different evidence to their released
oracles. OSWorld evaluates final virtual-machine and application state
with task-specific getters and metrics; WebArena and VisualWebArena
combine answer, URL, and programmatic state checks, with the latter also
requiring visual evidence; WorkArena checks task-specific ServiceNow
records and fields; and AssistantBench matches an open-web answer against
a reference answer. These mechanisms define the benchmark-side evidence
against which we compare the trajectory-level judgments.

\subsection{Labeling and Combination}
\label{sec:case-study}

We operationalize the Introduction's question---\textit{when should a CUA
verdict be distrusted?}---as a two-stage labeling decision. In Stage~1,
each recorded FAIL is labeled \textit{genuine agent failure} when the
trajectory does not satisfy the task and a human grader would also reject
it; \textit{evaluator false negative} when the observed outcome satisfies
the task but the checker rejects it; \textit{broken task} when the
specification is infeasible, the reference state is stale, or the
environment or harness blocks completion; or \textit{unclear} when the
released evidence cannot establish the correct judgment. In Stage~2,
only genuine failures receive a process diagnosis from the three-tier
taxonomy below. This sequence keeps benchmark faults from counting as
agent faults.

Two vision-enabled LLM annotators from different vendors---GPT-5.5
through OpenAI Codex CLI v0.144.5 and Anthropic Claude Sonnet---label all
150 trajectories independently and blind to each other's outputs. The
experiment artifacts retain the Claude model family but not its exact
Sonnet revision. Both receive the same written codebook and the same
step-level reasoning, actions, and screenshots. Human review includes
all 74 rows on which the LLMs disagree in verdict or category and a
seeded, benchmark-by-verdict stratified sample of 30 LLM-agreement rows.
Two human groups independently and blindly label the same 104 rows
through a screenshot-replay interface; thus, each reviewed trajectory
has four independent judgments. After annotation is complete, human
agreement determines the final label for reviewed rows, with
disagreements resolved by majority vote. The remaining 46 rows use the
two-LLM consensus because every LLM disagreement is human-reviewed.

\subsection{Three-Tier Diagnostic Taxonomy}

Once Stage~1 establishes that the agent genuinely failed, Stage~2
localizes the earliest decisive failure in the agent's
perception--action loop. We adapt MAST~\cite{cemri2025multi} from
multi-agent coordination to single-agent computer use, informed by
OSWorld's perception--action loop~\cite{osworld2024}. We remove intrinsically multi-agent
categories, reinterpret communication breakdowns as failures of state
propagation across steps, and elevate GUI grounding and long-horizon
verification. This separates failures in goal interpretation and
strategy from failures that prevent a plausible intent from being
executed or maintained~\cite{reflexion2023}. Appendix
\ref{app:failure-taxonomy-details} records the full mapping.

\paragraph{Tier 1---planning and specification.}
The three categories separate specification violations
(\textsc{T1-Spec}), repeated planning without a viable alternative
(\textsc{T1-Loop}), and plans built around nonexistent features or
capabilities (\textsc{T1-Hall}). These are intent or strategy failures before successful execution.

\paragraph{Tier 2---execution and grounding.}
Given a plausible intent, grounding failures (\textsc{T2-Ground}) select
the wrong coordinates or GUI control, state-propagation failures
(\textsc{T2-State}) lose context established earlier, and tool failures
(\textsc{T2-Tool}) use invalid arguments or encounter an action-space
limitation~\cite{computerUsingAgent2024,osworld2024}.

\paragraph{Tier 3---verification and feedback.}
The categories distinguish premature termination
(\textsc{T3-Term}), missed or incorrect verification
(\textsc{T3-Verif}), and feedback-blind repetition of actions that do
not change state (\textsc{T3-Noop}). These failures occur after an action
can be checked and corrected.

\paragraph{Boundary rule.}
We assign the earliest failure that makes the trajectory unsuccessful,
rather than the most visually salient later symptom. Repetition is
Tier~1 when the agent cannot develop an alternative strategy; it is
Tier~3 when the agent has a viable plan but fails to notice that an
action had no effect. The distinction is therefore causal rather than
visual.

\subsection{Results and Discussion}

\begin{table*}[t]
\centering
\scriptsize
\setlength{\tabcolsep}{3pt}
\renewcommand{\arraystretch}{1.02}
\begin{tabularx}{\textwidth}{@{}p{2.05cm} X rrrrrr@{}}
\toprule
\textbf{Benchmark} & \textbf{Released oracle and inspected evidence} &
\textbf{$n$} & \textbf{GF} & \textbf{EFN} & \textbf{Broken} &
\textbf{Unclear} & \textbf{Wrong (\%)} \\
\midrule
OSWorld & Execution-based getters over final VM/application state & 57 & 47 & 8 & 2 & 0 & 17.5 \\
WebArena & String, URL, and program checks on answer and DOM state & 23 & 18 & 5 & 0 & 0 & 21.7 \\
VisualWebArena & Visual and program checks on answer, URL, and DOM state & 23 & 18 & 3 & 0 & 2 & 13.0 \\
WorkArena & Programmatic checks over ServiceNow records and fields & 24 & 23 & 0 & 0 & 1 & 0.0 \\
AssistantBench & Open-web answer matching against a reference answer & 23 & 16 & 0 & 5 & 2 & 21.7 \\
\midrule
\textbf{All} & & \textbf{150} & \textbf{122} & \textbf{16} &
\textbf{7} & \textbf{5} & \textbf{15.3} \\
\bottomrule
\end{tabularx}

\vspace{2pt}

\begin{tabularx}{\textwidth}{@{}l X r X r X r rr@{}}
\toprule
\textbf{Tier} & \multicolumn{6}{l}{\textbf{Subcategory diagnoses among
genuine failures ($n=122$)}} & \textbf{Total} & \textbf{Share} \\
\cmidrule(lr){2-7}
& \textbf{Subcategory 1} & \textbf{$n$}
& \textbf{Subcategory 2} & \textbf{$n$}
& \textbf{Subcategory 3} & \textbf{$n$}
& \textbf{122} & \textbf{100\%} \\
\midrule
Tier 1 & Specification violation & 18 & Plan loop & 20 &
Hallucinated feature & 5 & 43 & 35.2\% \\
Tier 2 & Grounding & 9 & Tool arguments & 6 & State loss & 2 & 17 &
13.9\% \\
Tier 3 & Feedback-blind no-op & 36 & Premature stop & 8 & Missed
verification & 4 & 48 & 39.3\% \\
Other & \multicolumn{6}{l}{Genuine failure outside the three tiers} &
6 & 4.9\% \\
Ambiguous & \multicolumn{6}{l}{Released evidence does not distinguish
tiers after label combination} & 8 & 6.6\% \\
\bottomrule
\end{tabularx}
\caption{Unified audit setting and results. The upper panel combines
each benchmark's released oracle with its verdict audit; ``Wrong'' is
the percentage of audited FAILs that are evaluator false negatives or
broken tasks. The lower panel reports tier totals and the
non-overlapping Stage~2 subcategory diagnoses of genuine failures. GF
denotes genuine failure and EFN evaluator false negative.}
\label{tab:verdict-audit}
\end{table*}

\paragraph{Verdict reliability.}
Across the five benchmarks, \textbf{15.3\% of audited FAIL verdicts are
wrong} (95\% Wilson CI [10.4, 22.0]): 10.7\% [6.7, 16.6] are evaluator
false negatives and 4.7\% [2.3, 9.3] are broken tasks; another 3.3\%
[1.4, 7.6] remain unclear from the released evidence
(Table~\ref{tab:verdict-audit}). The benchmark mechanisms fail in
different ways. WebArena's answer matching rejects acceptable behavior
(21.7\% false negatives), while AssistantBench's open-web tasks are
vulnerable to environment drift, including dead search engines and
CAPTCHA walls (21.7\% broken tasks). OSWorld's state checkers miss
alternative valid solutions (14.0\%) and include a task that is
unsatisfiable on the released VM image. WorkArena's programmatic state
checkers produce no detected wrong verdicts in this sample (0/24).

\paragraph{Illustrative cases.}
WebArena scored FAIL after an agent correctly listed the five single-book
recommendations among the top-ten books posts. An AssistantBench eatery
query encountered a DuckDuckGo error and then a Google CAPTCHA.
OSWorld's battery-percentage task was impossible because
\texttt{/sys/class/power\_supply} was empty. These cases show why a
recorded failure should be distrusted when the trajectory demonstrates
task completion that the oracle does not recognize or when the task
cannot be completed in the released environment.

\paragraph{Failure diagnoses.}
Among the 122 genuine failures, the scalar score conceals a strongly
skewed diagnosis. Tier~3 verification and feedback failures account for
39.3\%. Feedback-blind no-op repetition alone accounts for 29.5\%, the
largest single category. Tier~1 planning failures account for 35.2\%,
including specification violations (14.8\%) and planning loops (16.4\%).
Tier~2 execution and grounding errors account for only 13.9\%; 6.6\% remain
ambiguous across tiers after label combination, and 4.9\% fall outside the
three tiers. For example, in a genuine WorkArena failure, the agent
clicked ``ALL RESULTS'' 27 times while ``No Results'' and the screenshot
remained unchanged, exemplifying the dominant Tier~3 no-op pattern.

\paragraph{Annotation reliability and observability.}
The two LLM judges agree substantially on Stage~1 verdicts
($\kappa=0.71$, raw agreement 92.7\%) but only moderately on Stage~2
diagnoses ($\kappa=0.41$). On the 104-row common blind-review set, the two
human groups reach $\kappa=0.59$ (85.6\% raw agreement); pairwise
human--LLM $\kappa$ ranges from 0.19 to 0.32 (76.0--81.7\% raw), and
four-rater Fleiss' $\kappa$ is 0.36. Because this set deliberately
contains every LLM disagreement, these are stress-set reliability
statistics rather than population agreement estimates.

We also quantify variability in the binary wrong-verdict score
(evaluator false negative or broken task versus all other Stage~1
outcomes) on these 104 commonly reviewed rows. Across GPT-5.5, Claude
Sonnet-5, human group~1, and human group~2, the annotator-specific rates
are 13.5\%, 7.7\%, 20.2\%, and 13.5\%, respectively: mean 13.7\%,
sample standard deviation 5.1 percentage points, sample variance 0.0026,
and range 7.7--20.2\%. The final 15.3\% estimate instead uses the
post-combination labels for all 150 rows and its Wilson interval above.

The evidence available to the judge materially changes these decisions:
removing screenshots flips 12 Codex verdicts and 13 Claude verdicts,
reduces the evaluator false negatives detected by the two judges from 14
to 8 and from 10 to 6, respectively, and lowers their Stage~1 agreement
to $\kappa=0.60$. Thus, screenshot-inclusive trajectories are necessary
both for auditing checker validity and for diagnosing genuine failures.

\paragraph{Implications.}
The audit directly answers the paper's motivating question: a CUA
failure verdict warrants distrust when replay evidence contradicts the
checker or reveals that the task or environment is invalid. Even when
the verdict is correct, it is incomplete unless the released trajectory
supports a diagnosis. The difference between substantial verdict
agreement and moderate diagnosis agreement also shows that fine-grained
labels require explicit operational rules and human calibration.
\begin{table*}[t]
\centering
{\footnotesize
\setlength{\tabcolsep}{4pt}
\renewcommand{\arraystretch}{1.1}
\renewcommand{\tabularxcolumn}[1]{m{#1}}
\begin{tabularx}{\linewidth}{@{}
  >{\raggedright\arraybackslash\hspace{0pt}}m{1.55cm}
  >{\hsize=1.18\hsize\linewidth=\hsize\raggedright\arraybackslash\hspace{0pt}}X
  >{\raggedright\arraybackslash\hspace{0pt}}m{2.45cm}
  >{\hsize=0.82\hsize\linewidth=\hsize\raggedright\arraybackslash\hspace{0pt}}X@{}}
\toprule
\textbf{Stage} & \textbf{Trustworthy baseline (\S\ref{sec:guidelines})} & \textbf{Threat addressed} & \textbf{Toward better benchmarks (\S\ref{sec:future})} \\
\midrule
Task construction & Replay a stratified subsample pre-release to confirm tasks are solvable as shipped; author tasks around workflows practitioners use; version tasks, environments, and oracles together, retiring stale ones & Broken or invalid tasks; drift; memorization & Living task pools and procedural refresh; expert-authored workflows and acceptance criteria \\
\midrule
Environment & Pin and disclose the execution stack---OS image, application versions, and scaffold packages (e.g., Playwright vs.\ PyAutoGUI); declare the retrieval boundary (\S\ref{sec:systemic-failures}); support deterministic replay & Irreproducible or scaffold-confounded scores; contamination & Seeded perturbations and self-evolving environments under deterministic replay \\
\midrule
Scoring & Check exactly what the task statement declares: accept every solution in that space and verify each constraint; audit checkers against valid alternatives and human labels; report false-negative rates & Evaluator false negatives; reward hacking & Calibrated judge- and rubric-based scoring of process and outcome \\
\midrule
Reporting & Report success rate plus a per-failure locus and process-stage diagnosis; release evidence-complete trajectories (screenshots, tool I/O, timestamps); disclose hardware, prices, and evaluation date & Diagnostic opacity; economically misleading comparisons & Automatic diagnostic probes; critical-window slack and deadline-miss curves (latency); dollar cost per verified success \\
\bottomrule
\end{tabularx}}
\caption{Benchmark-design guidelines by measurement stage
(Figure~\ref{fig:audit-framework}B). The second column lists the
baseline controls a benchmark needs for its verdicts to deserve trust
(\S\ref{sec:guidelines}); the last column previews the extensions of
\S\ref{sec:future}.}
\label{tab:guidelines}
\end{table*}

\section{Designing Reliable Benchmarks}
\label{sec:guidelines}

This section turns the audit results of \S\ref{sec:taxonomy} into the
baseline controls a CUA benchmark needs before its verdicts deserve
trust, following the four measurement stages of
Figure~\ref{fig:audit-framework}B; Table~\ref{tab:guidelines}
summarizes the controls for each stage.

\paragraph{Construct tasks that practitioners would recognize.}
Seven recorded failures (4.7\%) were broken tasks, not agent
failures: AssistantBench's open-web tasks hit dead search engines and
CAPTCHA walls, and one OSWorld task was unsatisfiable on the released
VM image. Pre-release replay and joint versioning of tasks,
environments, and oracles address this. Validity, however, demands
more than solvability: a task must exercise the workflow the
profession actually uses~\cite{dong2026cadworld}. Benchmark design
should thus involve domain experts in defining tasks and the artifacts
that count as professionally acceptable.

\paragraph{Disclose the execution stack and retrieval boundary.}
An agent's score is conditioned on its scaffold: the same model can
look competent under one scaffold and helpless under another because
the action and observation spaces differ, not the capability. Scores are therefore comparable only
when the OS image, application versions, and scaffold packages are
pinned and disclosed, and when released environments replay
deterministically. The retrieval boundary needs the same rigor. A
blanket network ban is inappropriate when live lookup is part of the
task; instead, benchmarks should declare a policy matched to the STC
threat model of \S\ref{sec:systemic-failures}: whitelist network access
in closed worlds, timestamp tasks and log retrieved sources on the live
web, and audit retrieval indexes and evaluation-log reuse in enterprise
settings~\cite{searchTimeContam2025}.

\paragraph{Score the task that was stated.}
A checker is valid when it tests exactly what the task statement
declares (no less or more). Testing less rejects legitimate work: 16
of 150 audited failures (10.7\%) were evaluator false negatives,
including five WebArena answers rejected by string matching and eight
OSWorld solutions whose alternative valid paths the state checkers
missed. Well-designed oracles therefore accept any solution inside the
declared space, using type-aware or semantic normalization whenever the
answer space is open~\cite{webarenaVerifiedGithub,agentAsJudge2024}.
Testing more loosely than the statement invites hacking in the other
direction: when a task constrains the method, an outcome-only
check would credit an equivalent artifact built with a different
workbench, rewarding a capability the task never claimed to measure.
Task statements and checkers must be co-designed so this ``exactly''
is tested with both positive and negative cases.

\paragraph{Report evidence, not just a verdict.}
Success rate remains the right top-line metric, but a scalar cannot
direct engineering effort. Each failure should carry a two-level
diagnosis: first its locus (task/environment, trajectory/evidence,
evaluator, or agent) and then, for genuine agent failures, the process
stage and reason category. In our audit, verification and feedback
failures (39.3\%) and planning failures (35.2\%) dominated execution
and grounding errors (13.9\%)---a distribution the scalar conceals
entirely. Diagnosis also depends on
the released evidence: removing screenshots flipped 12 Codex and 13
Claude verdicts and roughly halved the evaluator false negatives either
judge detected. Benchmarks should therefore release actions,
observations, screenshots, tool I/O, timestamps, and state transitions
alongside the scores~\cite{cemri2025multi,selfEvolvingBench2025}, and
disclose serving hardware, price schedules, and the evaluation date so
that latency and cost can be reconstructed; \S\ref{sec:future} develops
the corresponding first-class metrics.

\section{Design Better Benchmarks}
\label{sec:future}

This section develops, for each measurement stage of
\S\ref{sec:guidelines}, methods that are more diagnostic, scalable,
and harder to saturate, as previewed in Table~\ref{tab:guidelines}.

\paragraph{Task construction: living pools and professional standards.}
Static task sets saturate and leak, so tasks must evolve faster than
the models they test: Agents' Last Exam maintains a living task pool on
a declared schedule, and self-evolving or procedural task generation
varies file names, layouts, and data values to defeat memorization
~\cite{agentsLastExam2026,agentgym2024,selfEvolvingBench2025,wang2019paired}.
The second frontier is who defines the tasks. Existing suites measure
whether an agent can operate professional software; almost none ask
whether the profession would accept the result. Extending the
domain-validity guideline of \S\ref{sec:guidelines}, domain experts
should author not only the workflows but also the acceptance
criteria for finished, correct work~\cite{prosoftarena2025,dong2026cadworld}.

\paragraph{Environment: realistic variation under deterministic replay.}
Beyond a pinned, disclosed stack, better environments inject controlled
variation: seeded randomization reproduces real-world mess like network
lag, transient UI drift, and asynchronous updates, testing robustness
without giving up deterministic replay, while self-evolving
environments such as AgentGym grow alongside the agent so the
evaluation never hits a ceiling
~\cite{wang2019paired,agentgym2024,selfEvolvingBench2025}.

\paragraph{Scoring: calibrated judges over process, not just outcome.}
As open-ended tasks outgrow brittle scripts, scoring is shifting from
outcome-only checks to rubric- and process-aware instruments: WeaveBench
inspects deliverables together with screenshots, files, and action
traces; Odysseys grades rubric completion per step; and CUARewardBench
separately tests outcome and process reward models
~\cite{weavebench2026,odysseys2026,cuaRewardBench2026}. The instrument
that makes this scale is \textit{Agent-as-a-Judge}: a judge more capable
than the system under test, performing verification against logged
trajectories. This judge is trusted only after meta-evaluation against human labels, as
\S\ref{sec:case-study} practices
~\cite{agentAsJudge2024,agentRewardBench2025}. With
practitioner-authored rubrics, the same machinery can grade work by
professional standards rather than generic step checklists.

\paragraph{Reporting: automatic diagnostics before expensive judges.}
The diagnostic reporting of \S\ref{sec:guidelines} does not require
annotating every failure with humans or judges. The most common failure
in our audit that involves an action repeated while the screen observably does not
change (29.5\% of genuine failures), can be flagged by scanning each
trajectory for repeated actions with unchanged consecutive
observations, at zero model cost; judge or human effort is then spent
only on the planning and grounding failures that need semantic
judgment. Such a probe must first be validated against human labels,
since legitimate no-change intervals, such as page loads and
asynchronous rendering, would otherwise be misflagged.

\paragraph{Reporting: latency and cost as first-class metrics.}
Standard reports say whether an agent succeeded, not whether it acted
in time or at what price. Short-lived controls such as BIOS prompts and
countdown links can disappear before an agent acts. For such a task, let $R$ be
the number of benchmark-controlled observation--action opportunities
during which the target remains actionable and $r\in\{1,\ldots,R\}$ the
opportunity on which the correct action is issued;
\textit{critical-window slack}, $S_{\mathrm{win}}=(R-r+1)/R$ (zero on a
miss), is hardware-independent because it counts environment
opportunities, not seconds. Physical latency cannot be
normalized away, so wall-clock deadline-miss rate and
success-versus-deadline curves at declared compute limits and hardware
complete the picture. Token counts are not comparable across models or
providers; with $C$ covering agent inference, tool calls, environment
runtime, and verification, reports should give dollar cost per attempt
($C/N_{\mathrm{attempt}}$) and per verified success
($C/N_{\mathrm{success}}$) with the price schedule, exposing a
cost--success frontier that trajectory pruning makes cheaper to occupy:
AgentDiet cuts input tokens by up to 59.7\% without sacrificing
success~\cite{agentdiet2025}.

\section{Conclusion}

We showed how contamination, evaluator fragility, and diagnostic
opacity distort CUA benchmark verdicts. We hope readers see scores as
measurement-pipeline outputs, judge when a verdict deserves trust, and
build benchmarks that earn it.

\section*{Limitations}

\paragraph{Audit scope.} Our claims are restricted to GUI computer-use evaluation. The audit covers 150 trajectories from five benchmarks, all drawn from public artifacts in web, enterprise-workflow, and desktop-control settings. The specialized healthcare, CAD/EDA, scientific-instrument, and creative-production resources listed in Table~\ref{tab:cua-evaluation-landscape} are contextual and receive no empirical claim. Our 2026 benchmark discussion establishes continuity with newer long-horizon evaluation designs but does not retrospectively add them to the audit sample. Per-benchmark samples are small (23--57), so the benchmark-level rates in Table~\ref{tab:verdict-audit} carry wide intervals and should be read as evidence that evaluator error occurs at a material rate, not as calibrated per-benchmark estimates. We sampled only failure-scored trajectories, so we estimate evaluator false negatives but say nothing about false positives; the complementary audit of PASS verdicts remains open.

\paragraph{Survey scoping.} Screening was iterative and interleaved with drafting rather than executed as a single logged pass, so we report the included corpus rather than PRISMA-style identification and exclusion counts (Appendix~\ref{app:scoping-protocol}).

\paragraph{Unvalidated design inference.} The detection-cost argument of \S\ref{sec:future} is conceptual; we do not validate a programmatic probe against our human labels, and we make no claim about what accuracy such a probe would achieve.

\bibliography{references}

@String{Computer = "{IEEE} Computer" }

@inproceedings{surveyAgents2025,
  title={Evaluation and benchmarking of llm agents: A survey},
  author={Mohammadi, Mahmoud and Li, Yipeng and Lo, Jane and Yip, Wendy},
  booktitle={Proceedings of the 31st ACM SIGKDD Conference on Knowledge Discovery and Data Mining V. 2},
  pages={6129--6139},
  year={2025}
}

@inproceedings{webarenaVerifiedGithub,
    title={WebArena Verified: Reliable Evaluation for Web Agents},
    author={Amine {El hattami} and Megh Thakkar and Nicolas Chapados and Christopher Pal},
    booktitle={Workshop on Scaling Environments for Agents},
    year={2025},
    url={https://openreview.net/forum?id=94tlGxmqkN}
}

@misc{visualwebarenaRepo,
  title        = {{VisualWebArena}: A Benchmark for Multimodal Agents},
  howpublished = {GitHub repository},
  year         = {2024},
  key          = {visualwebarenaRepo},
  note         = {Accessed 2026-01-27},
}

@misc{workarenaRepo,
      title={WorkArena++: Towards Compositional Planning and Reasoning-based Common Knowledge Work Tasks}, 
      author={Léo Boisvert and Megh Thakkar and Maxime Gasse and Massimo Caccia and Thibault Le Sellier De Chezelles and Quentin Cappart and Nicolas Chapados and Alexandre Lacoste and Alexandre Drouin},
      year={2024},
      eprint={2407.05291},
      archivePrefix={arXiv},
      primaryClass={cs.AI},
      url={https://arxiv.org/abs/2407.05291}, 
}

@article{workarena2024,
  title={Workarena: How capable are web agents at solving common knowledge work tasks?},
  author={Drouin, Alexandre and Gasse, Maxime and Caccia, Massimo and Laradji, Issam H and Del Verme, Manuel and Marty, Tom and Boisvert, L{\'e}o and Thakkar, Megh and Cappart, Quentin and Vazquez, David and others},
  journal={arXiv preprint arXiv:2403.07718},
  year={2024}
}

@article{osworld2024,
  title={Osworld: Benchmarking multimodal agents for open-ended tasks in real computer environments},
  author={Xie, Tianbao and Zhang, Danyang and Chen, Jixuan and Li, Xiaochuan and Zhao, Siheng and Cao, Ruisheng and Hua, Toh J and Cheng, Zhoujun and Shin, Dongchan and Lei, Fangyu and others},
  journal={Advances in Neural Information Processing Systems},
  volume={37},
  pages={52040--52094},
  year={2024}
}

@misc{computerUsingAgent2024,
  title        = {Computer-Using Agent},
  author       = {{OpenAI}},
  howpublished = {OpenAI blog},
  year         = {2025},
  month        = jan,
  key          = {computerUsingAgent2024},
  note         = {Published 2025-01-23; accessed 2026-01-27},
}

@article{searchTimeContam2025,
  title        = {Search-Time Data Contamination},
  author       = {Han, Ziwen and Mankikar, Meher and Michael, Julian and Wang, Zifan},
  journal      = {arXiv preprint arXiv:2508.13180},
  year         = {2025},
}

@article{agentAsJudge2024,
  title        = {Agent-as-a-Judge: Evaluate Agents with Agents},
  author       = {Zhuge, Mingchen and Zhao, Changsheng and Ashley, Dylan and Wang, Wenyi and Khizbullin, Dmitrii and Xiong, Yunyang and Liu, Zechun and Chang, Ernie and Krishnamoorthi, Raghuraman and Tian, Yuandong and Shi, Yangyang and Chandra, Vikas and Schmidhuber, J{\"u}rgen},
  journal      = {arXiv preprint arXiv:2410.10934},
  year         = {2024},
}

@article{agentRewardBench2025,
  title={Agentrewardbench: Evaluating automatic evaluations of web agent trajectories},
  author={L{\`u}, Xing Han and Kazemnejad, Amirhossein and Meade, Nicholas and Patel, Arkil and Shin, Dongchan and Zambrano, Alejandra and Sta{\'n}czak, Karolina and Shaw, Peter and Pal, Christopher J and Reddy, Siva},
  journal={arXiv preprint arXiv:2504.08942},
  year={2025}
}

@article{selfEvolvingBench2025,
  title={Towards Self-Evolving Benchmarks: Synthesizing Agent Trajectories via Test-Time Exploration under Validate-by-Reproduce Paradigm},
  author={Guo, Dadi and Zhou, Tianyi and Liu, Dongrui and Qian, Chen and Ren, Qihan and Shao, Shuai and Fan, Zhiyuan and Fung, Yi R and Wang, Kun and Zhang, Linfeng and others},
  journal={arXiv preprint arXiv:2510.00415},
  year={2025}
}

@article{agentdiet2025,
  title={Improving the efficiency of LLM agent systems through trajectory reduction},
  author={Xiao, Yuan-An and Gao, Pengfei and Peng, Chao and Xiong, Yingfei},
  journal={arXiv preprint arXiv:2509.23586},
  year={2025}
}

@inproceedings{agentgym2024,
  title        = {{AgentGym}: Evaluating and Training Large Language Model-based Agents across Diverse Environments},
  author       = {Xi, Zhiheng and Ding, Yiwen and others},
  booktitle    = {Proceedings of the 63rd Annual Meeting of the Association for Computational Linguistics (ACL)},
  pages        = {27914--27961},
  year         = {2025},
}

@inproceedings{androidworld2024,
  title        = {AndroidWorld: A Dynamic Benchmarking Environment for Autonomous Agents},
  author       = {Rawles, Christopher and others},
  booktitle    = {Advances in Neural Information Processing Systems (NeurIPS)},
  year         = {2024},
}

@inproceedings{appagent2023,
  title        = {AppAgent: Multimodal Agents as Smartphone Users},
  author       = {Zhang, Chi and others},
  booktitle    = {Proceedings of the IEEE/CVF Conference on Computer Vision and Pattern Recognition (CVPR)},
  year         = {2024},
}

@misc{mobileworld2025,
  title        = {MobileWorld: Benchmarking Autonomous Mobile Agents in Agent-User Interactive, and {MCP}-Augmented Environments},
  author       = {Kong, Quyu and others},
  howpublished = {arXiv},
  year         = {2025},
  note         = {arXiv:2512.19432},
}

@article{wang2024mobile,
  title={Mobile-agent-v2: Mobile device operation assistant with effective navigation via multi-agent collaboration},
  author={Wang, Junyang and Xu, Haiyang and Jia, Haitao and Zhang, Xi and Yan, Ming and Shen, Weizhou and Zhang, Ji and Huang, Fei and Sang, Jitao},
  journal={Advances in Neural Information Processing Systems},
  volume={37},
  pages={2686--2710},
  year={2024}
}

@article{mobileenv2023,
  title={Mobile-Env: Building qualified evaluation benchmarks for {LLM-GUI} interaction},
  author={Zhang, Danyang and others},
  journal={arXiv preprint arXiv:2305.08144},
  year={2023}
}

@inproceedings{reflexion2023,
  title        = {Reflexion: Language Agents with Verbal Reinforcement Learning},
  author       = {Shinn, Noah and Cassano, Federico and Gopinath, Ashwin and Narasimhan, Karthik and Yao, Shunyu},
  booktitle    = {Advances in Neural Information Processing Systems (NeurIPS)},
  year         = {2023},
}

@inproceedings{webarena2023,
  title        = {WebArena: A Realistic Web Environment for Building Autonomous Agents},
  author       = {Zhou, Shuyan and others},
  booktitle    = {Advances in Neural Information Processing Systems (NeurIPS)},
  year         = {2023},
}

@misc{qwenAgentic2026,
  title        = {Alibaba's Qwen App Advances Agentic AI Strategy by Turning Core Ecosystem Services into Executable AI Capabilities},
  author       = {{Alibaba Group}},
  howpublished = {Alibaba Group announcement},
  year         = {2026},
  month        = jan,
  key          = {qwenAgentic2026},
  note         = {Accessed 2026-01-31},
}

@misc{doubaoPhoneAssistant2025,
  title        = {ByteDance's agentic {AI} smartphone dials up a digital backlash from China's top apps},
  howpublished = {South China Morning Post},
  year         = {2025},
  month        = dec,
  author       = {Eunice Xu},
  key          = {doubaoPhoneAssistant2025},
  note         = {Published 2025-12-07; accessed 2026-01-31},
}

@misc{openaiOperator2025,
  title        = {OpenAI Operator agent can automate tasks such as vacation planning},
  howpublished = {CNBC},
  year         = {2025},
  month        = jan,
  author       = {Hayden Field},
  key          = {openaiOperator2025},
  note         = {Published 2025-01-23; accessed 2026-01-31},
}

@misc{googleAiModeAgentic2025,
  title        = {{AI} Mode in Search gets new agentic features and expands globally},
  author       = {Stein, Robby},
  howpublished = {Google blog (Products / Search)},
  year         = {2025},
  month        = aug,
  key          = {googleAiModeAgentic2025},
  note         = {Published 2025-08-21; accessed 2026-01-31},
}

@inproceedings{mind2web2023,
  title        = {Mind2Web: Towards a Generalist Agent for the Web},
  author       = {Deng, Xiang and Gu, Yu and Zheng, Boyuan and Chen, Shijie and Stevens, Samuel and Wang, Xuehai and Sun, Huan and Su, Yu},
  booktitle    = {Advances in Neural Information Processing Systems (NeurIPS) Datasets and Benchmarks Track},
  year         = {2023},
}

@article{browsergym2024,
  title        = {The {BrowserGym} Ecosystem for Web Agent Research},
  author       = {Le Sellier de Chezelles, Thibault and Gasse, Maxime and Lacoste, Alexandre and Drouin, Alexandre and Caccia, Massimo and Boisvert, Léo and others},
  journal      = {Transactions on Machine Learning Research (TMLR)},
  year         = {2024},
}

@inproceedings{weblinx2024,
  title        = {{WebLINX}: Real-World Website Navigation with Multi-Turn Dialogue},
  author       = {Lù, Xing Han and Kasner, Zdeněk and Reddy, Siva},
  booktitle    = {Proceedings of the IEEE/CVF Conference on Computer Vision and Pattern Recognition (CVPR)},
  year         = {2024},
}

@inproceedings{miniwobplusplus2018,
  title     = {Reinforcement Learning on Web Interfaces using Workflow-Guided Exploration},
  author    = {Liu, Evan Zheran and Guu, Kelvin and Pasupat, Panupong and Shi, Tianlin and Liang, Percy},
  booktitle = {International Conference on Learning Representations (ICLR)},
  year      = {2018},
  note      = {arXiv:1802.08802},
}

@inproceedings{visualwebarena2024,
  title        = {VisualWebArena: Evaluating Multimodal Agents on Realistic Visual Web Tasks},
  author       = {Koh, Jing Yu and others},
  booktitle    = {Proceedings of the 62nd Annual Meeting of the Association for Computational Linguistics (ACL)},
  year         = {2024},
}

@inproceedings{webvoyager2024,
  title        = {WebVoyager: Building an End-to-End Web Agent with Large Multimodal Models},
  author       = {He, Shijie and others},
  booktitle    = {Proceedings of the 62nd Annual Meeting of the Association for Computational Linguistics (ACL)},
  year         = {2024},
}

@article{cemri2025multi,
  title={Why do multi-agent llm systems fail?},
  author={Cemri, Mert and Pan, Melissa Z and Yang, Shuyi and Agrawal, Lakshya A and Chopra, Bhavya and Tiwari, Rishabh and Keutzer, Kurt and Parameswaran, Aditya and Klein, Dan and Ramchandran, Kannan and others},
  journal={arXiv preprint arXiv:2503.13657},
  year={2025}
}

@inproceedings{bavaresco2025llms, title={Llms instead of human judges? a large scale empirical study across 20 nlp evaluation tasks}, author={Bavaresco, Anna and Bernardi, Raffaella and Bertolazzi, Leonardo and Elliott, Desmond and Fern{\'a}ndez, Raquel and Gatt, Albert and Ghaleb, Esam and Giulianelli, Mario and Hanna, Michael and Koller, Alexander and others}, booktitle={Proceedings of the 63rd Annual Meeting of the Association for Computational Linguistics (Volume 2: Short Papers)}, pages={238--255}, year={2025} }

@article{wang2019paired,
  title={Paired open-ended trailblazer (poet): Endlessly generating increasingly complex and diverse learning environments and their solutions},
  author={Wang, Rui and Lehman, Joel and Clune, Jeff and Stanley, Kenneth O},
  journal={arXiv preprint arXiv:1901.01753},
  year={2019}
}

@article{yehudai2025survey,
  title={Survey on Evaluation of {LLM}-based Agents},
  author={Yehudai, Asaf and Eden, Lilach and Li, Alan and Uziel, Guy and Zhao, Yilun and Bar-Haim, Roy and Cohan, Arman and Shmueli-Scheuer, Michal},
  journal={arXiv preprint arXiv:2503.16416},
  year={2025}
}

@article{osworld2_2026,
  title={{OSWorld 2.0}: Benchmarking Computer Use Agents on Long-Horizon Real-World Tasks},
  author={Yuan, Mengqi and Zhou, Zilong and Xiong, Xinzhuang and Wu, Weiming and Sun, Jiayang and Song, Jiamin and others},
  journal={arXiv preprint arXiv:2606.29537},
  year={2026},
  url={https://arxiv.org/abs/2606.29537}
}

@article{agentsLastExam2026,
  title={Agents' Last Exam},
  author={Sun, Yiyou and Han, Xinyang and Zhang, Weichen and Pang, Yuanbo and Wang, Tianyu and Cao, Yuhan and Huang, Yixiao and others},
  journal={arXiv preprint arXiv:2606.05405},
  year={2026},
  url={https://arxiv.org/abs/2606.05405}
}

@article{odysseys2026,
  title={Odysseys: Benchmarking Web Agents on Realistic Long Horizon Tasks},
  author={Jang, Lawrence Keunho and Koh, Jing Yu and Fried, Daniel and Salakhutdinov, Ruslan},
  journal={arXiv preprint arXiv:2604.24964},
  year={2026},
  url={https://arxiv.org/abs/2604.24964}
}

@article{weavebench2026,
  title={WeaveBench: A Long-Horizon, Real-World Benchmark for Computer-Use Agents with Hybrid Interfaces},
  author={Li, Wanli and Zhou, Bowen and Yu, Yunyao and Xu, Zhou and Yang, Yifan and Li, Dongsheng and Shan, Caihua},
  journal={arXiv preprint arXiv:2606.09426},
  year={2026},
  url={https://arxiv.org/abs/2606.09426}
}

@inproceedings{cuaRewardBench2026,
  title={{CUARewardBench}: A Benchmark for Evaluating Reward Models for Computer-Using Agents},
  author={Lin, Haojia and Tan, Xiaoyu and Qin, Yulei and Xu, Zihan and Shi, Yuchen and Li, Zongyi and Li, Gang and Cai, Shaofei and Cai, Siqi and Fu, Chaoyou and others},
  booktitle={Proceedings of the 43rd International Conference on Machine Learning},
  year={2026},
  url={https://arxiv.org/abs/2510.18596}
}

@article{healthadminbench2026,
  title={{HealthAdminBench}: Evaluating Computer-Use Agents on Healthcare Administration Tasks},
  author={Bedi, Suhana and Welch, Ryan and Steinberg, Ethan and Wornow, Michael and Kim, Taeil Matthew and Ahmed, Haroun and Sterling, Peter and Purohit, Bravim and Akram, Qurat and Acosta, Angelic and Nubla, Esther and Sharma, Pritika and Pfeffer, Michael A. and Koyejo, Sanmi and Shah, Nigam H.},
  journal={arXiv preprint arXiv:2604.09937},
  year={2026},
  url={https://arxiv.org/abs/2604.09937}
}

@article{medcuabench2026,
  title={{MedCUA-Bench}: A Screenshot-Only Benchmark for Clinical Computer-Use Agents},
  author={Yu, Jia and Wang, Zilong and Jiang, Xinyang and Li, Dongsheng and Wang, Shuo},
  journal={arXiv preprint arXiv:2606.03203},
  year={2026},
  url={https://arxiv.org/abs/2606.03203}
}

@article{guieda2025,
  title={Using {GUI} Agent for Electronic Design Automation},
  author={Li, Chunyi and Li, Longfei and Zhang, Zicheng and Liu, Xiaohong and Tang, Min and Lin, Weisi and Zhai, Guangtao},
  journal={arXiv preprint arXiv:2512.11611},
  year={2025},
  url={https://arxiv.org/abs/2512.11611}
}

@article{labosbench2026,
  title={{LabOSBench}: Benchmarking Computer Use Agents for Scientific Instrument Control},
  author={Zou, Anqi and Deng, Han and Zhang, Chengyu and Hu, Junquan and Wang, Yu and Xing, Yuxiang and Zhang, Aokai and Zhang, Hanling and Liu, Zhaoyang and Fei, Ben and Wang, Zhihui and Ouyang, Wanli},
  journal={arXiv preprint arXiv:2606.16802},
  year={2026},
  url={https://arxiv.org/abs/2606.16802}
}

@article{deskcraft2026,
  title={{DeskCraft}: Benchmarking Desktop Agents on Professional Workflows and Human-in-the-Loop Collaboration},
  author={Wang, Wenkai and Xiong, Tao and Ni, Jingchen and Bao, Yunpeng and Li, Xiyun and Liu, Tianqi and Guo, Hongcan and Huang, Zilong and Zhang, Shengyu},
  journal={arXiv preprint arXiv:2606.03103},
  year={2026},
  url={https://arxiv.org/abs/2606.03103}
}

@inproceedings{screenspotpro2025,
  title={{ScreenSpot-Pro}: {GUI} Grounding for Professional High-Resolution Computer Use},
  author={Li, Kaixin and Meng, Ziyang and Lin, Hongzhan and Luo, Ziyang and Tian, Yuchen and Ma, Jing and Huang, Zhiyong and Chua, Tat-Seng},
  booktitle={Proceedings of the 33rd ACM International Conference on Multimedia},
  pages={8778--8786},
  year={2025},
  doi={10.1145/3746027.3755688},
  url={https://arxiv.org/abs/2504.07981}
}

@article{prosoftarena2025,
  title={{ProSoftArena}: Benchmarking Hierarchical Capabilities of Multimodal Agents in Professional Software Environments},
  author={Ai, Jiaxin and Feng, Yukang and Zhang, Fanrui and Sun, Jianwen and Li, Zizhen and Li, Chuanhao and Chang, Yifan and Wu, Wenxiao and Wang, Ruoxi and Zhai, Mingliang and Zhang, Kaipeng},
  journal={arXiv preprint arXiv:2601.02399},
  year={2025},
  url={https://arxiv.org/abs/2601.02399}
}

@article{guivscli2026,
  title={{GUI} vs.\ {CLI}: Execution Bottlenecks in Screen-Only and Skill-Mediated Computer-Use Agents},
  author={Zhou, Xiao and Zhang, Siyue and Zhao, Yilun and Wei, Jinbiao and Song, Tingyu and Cohan, Arman and Zhao, Chen},
  journal={arXiv preprint arXiv:2606.24551},
  year={2026},
  url={https://arxiv.org/abs/2606.24551}
}

@inproceedings{scienceboard2026,
  title={{ScienceBoard}: Evaluating Multimodal Autonomous Agents in Realistic Scientific Workflows},
  author={Sun, Qiushi and Liu, Zhoumianze and Ma, Chang and Ding, Zichen and Xu, Fangzhi and Yin, Zhangyue and Zhao, Haiteng and Wu, Zhenyu and Cheng, Kanzhi and Liu, Zhaoyang and Wang, Jianing and Li, Qintong and Tang, Xiangru and Xie, Tianbao and Feng, Xiachong and Li, Xiang and Kao, Ben and Wang, Wenhai and Qi, Biqing and Kong, Lingpeng and Wu, Zhiyong},
  booktitle={The Fourteenth International Conference on Learning Representations},
  year={2026},
  url={https://arxiv.org/abs/2505.19897}
}

@inproceedings{spider2v2024,
  title={{Spider2-V}: How Far Are Multimodal Agents From Automating Data Science and Engineering Workflows?},
  author={Cao, Ruisheng and Lei, Fangyu and Wu, Haoyuan and Chen, Jixuan and Fu, Yeqiao and Gao, Hongcheng and Xiong, Xinzhuang and Zhang, Hanchong and Mao, Yuchen and Hu, Wenjing and Xie, Tianbao and Xu, Hongshen and Zhang, Danyang and Wang, Sida and Sun, Ruoxi and Yin, Pengcheng and Xiong, Caiming and Ni, Ansong and Liu, Qian and Zhong, Victor and Chen, Lu and Yu, Kai and Yu, Tao},
  booktitle={Advances in Neural Information Processing Systems},
  volume={37},
  pages={107703--107744},
  year={2024},
  url={https://arxiv.org/abs/2407.10956}
}

@inproceedings{psbench2026,
  title={{PSBench}: Editing Image via {GUI} Agents in Photoshop},
  author={Zhang, Yinuo and Cheng, Zian and Zhao, Ziya and Li, Zongyu and Liu, Bingshuo and Liu, Qingbin and Cai, Junxian and Chen, Xi and Tu, Zhiying and Chu, Dianhui and Yu, Xiaoyan and Sui, Dianbo},
  booktitle={Proceedings of the 43rd International Conference on Machine Learning},
  series={Proceedings of Machine Learning Research},
  volume={306},
  year={2026},
  url={https://openreview.net/forum?id=DWATchaS4U}
}

@misc{dong2026cadworld,
  title={{CADWorld}: A {CAD}-Centric Benchmark for Spatial, Precise, and Long-Horizon Computer-Use Agents},
  author={Dong, Zihan and Liu, Yuanzhe and Ma, Zhiyuan and Li, Kaixin and Zhan, Qishi},
  year={2026},
  note={Manuscript},
  url={https://github.com/Zdong104/CADWORLD}
}

\clearpage
\appendix

\section{Benchmark Details}
\label{app:domain-details}

This appendix gives supporting detail for the GUI-CUA benchmarks summarized
in Table~\ref{tab:cua-evaluation-landscape}.

\subsection{Specialized Professional GUI Use}

\paragraph{Healthcare.}
HealthAdminBench evaluates end-to-end prior-authorization, appeals, and
equipment-order workflows across an EHR, payer portals, and a fax system,
using fine-grained deterministic checkpoints in addition to task
success~\cite{healthadminbench2026}. MedCUA-Bench instead reconstructs
clinical software from real manuals and open-source systems, separating
intent- and step-level goals and checking both completion and clinical
safety dimensions~\cite{medcuabench2026}. ProSoftArena adds a broader
Health \& Medicine split covering GUI operation of ImageJ, ChemDraw, and
RGui for diagnostic, pharmaceutical, and public-health work
~\cite{prosoftarena2025}.

\paragraph{CAD and electronic design.}
GUI-EDA targets screenshot-grounded interaction with five industrial
electronic-design CAD tools and reports comprehension and execution
accuracy~\cite{guieda2025}. CADWorld targets spatially precise,
long-horizon GUI interaction in CAD software~\cite{dong2026cadworld}.
ProSoftArena contributes execution-checked
AutoCAD and SolidWorks tasks~\cite{prosoftarena2025}; the matched
GUI-vs-CLI benchmark contributes a screen-only GUI arm that includes
FreeCAD artifact-editing tasks~\cite{guivscli2026}.

\paragraph{Scientific workflows.}
LabOSBench evaluates GUI agents on eight simulated scientific instruments,
with execution-based checks at subtask and end-to-end
levels~\cite{labosbench2026}. ScienceBoard covers real scientific software
and 169 workflows across six disciplines, including GUI-only and
GUI--CLI tasks~\cite{scienceboard2026}. Spider2-V evaluates GUI management
of professional data-science and engineering systems across the data
pipeline~\cite{spider2v2024}. ProSoftArena supplies further ImageJ,
ChemDraw, ArcGIS, ANSYS, MultiSim, and RGui tasks
~\cite{prosoftarena2025}.

\paragraph{Creative production.}
DeskCraft covers long-horizon design, video, audio, and 3D-creation
workflows and additionally evaluates mid-task clarification and post-task
revision~\cite{deskcraft2026}. PSBench contributes 600 human-annotated
Photoshop tasks and category-specific checks for non-destructive image
editing~\cite{psbench2026}; ProSoftArena adds execution-checked Photoshop
and Illustrator tasks~\cite{prosoftarena2025}.
ScreenSpot-Pro supplies a complementary, static test of target localization
in high-resolution professional interfaces; because it measures grounding
rather than completed workflows, we treat it as an additional capability
evaluation rather than an end-to-end application benchmark~\cite{screenspotpro2025}.

\subsection{Web Navigation}

\paragraph{Semantic web-navigation benchmarks.}
\begin{itemize}
  \item \textbf{WebArena:} A foundational, self-hosted benchmark involving functional e-commerce, forums, and GitLab instances. It shifted the focus from static snapshots to \textit{functional correctness} in stateful environments~\cite{webarena2023,webarenaVerifiedGithub}.
  \item \textbf{Mind2Web:} A large-scale dataset of real-world web trajectories across $>130$ domains, highlighting the difficulty of generalizing across diverse UI patterns~\cite{mind2web2023}.
  \item \textbf{MiniWoB++:} A reinforcement learning-style environment with controlled, programmatic tasks that serve as a baseline for low-level browser interactions~\cite{miniwobplusplus2018}.
  \item \textbf{AgentRewardBench:} A meta-benchmark that evaluates the reliability of reward models (evaluators) themselves, revealing that heuristic evaluators in web tasks often have high false negative rates~\cite{agentRewardBench2025}.
\end{itemize}

\paragraph{Multimodal and enterprise web benchmarks.}
\begin{itemize}
  \item \textbf{VisualWebArena:} Requires agents to process visual cues (e.g., ``Find the red dress in the image''), exposing the limitations of agents that cannot ``see'' beyond DOM structure~\cite{visualwebarenaRepo,visualwebarena2024}.
  \item \textbf{WebVoyager:} An end-to-end framework for evaluating agents on the live web, highlighting divergence between snapshot-based evaluation and real-world performance drift under dynamic content updates~\cite{webvoyager2024}.
  \item \textbf{WorkArena:} Focuses on specialized enterprise workflows (ServiceNow), testing the limits of agents in administrative environments where state mutation is complex and requires strict adherence to business logic~\cite{workarena2024,workarenaRepo}.
  \item \textbf{WebLINX:} A multimodal benchmark for mid-horizon web navigation that emphasizes alignment between visual frames and HTML structures~\cite{weblinx2024}.
\end{itemize}

\subsection{OS and Mobile}

\paragraph{Long-horizon professional computer use.}
\begin{itemize}
  \item \textbf{OSWorld~2.0} extends desktop evaluation to 108 end-to-end workflows whose median human completion time is about 1.6 hours. It reports both binary and partial completion and analyzes failures involving constraint tracking, intermediate information, hidden state, and verification~\cite{osworld2_2026}.
  \item \textbf{Agents' Last Exam (ALE)} contains 1K+ expert-sourced tasks spanning 55 professional subfields and 13 industry clusters. Agents operate real software in reproducible Windows and Linux sandboxes through GUI and terminal interfaces; deterministic and judge-based graders compare output artifacts against hidden references~\cite{agentsLastExam2026}.
  \item \textbf{WeaveBench} contains 114 hybrid-interface tasks that combine GUI, CLI, code, and browser operations. Its trajectory-aware judge inspects files, screenshots, logs, and action traces, providing a contemporary example of evidence-rich evaluation~\cite{weavebench2026}.
\end{itemize}

\paragraph{Visual grounding systems.}
\noindent\textbf{OSWorld:} Evaluates multimodal agents in real computer environments spanning Ubuntu, Windows, and macOS, with 369 tasks requiring cross-application workflows~\cite{osworld2024}.

\noindent\textbf{AppAgent:} An architecture for multimodal agents as smartphone users that illustrates how even conceptually correct identification of a UI element can fail due to small spatial grounding errors~\cite{appagent2023}.

\noindent\textbf{OpenAI Computer-Using Agent:} A frontier system that operates directly on the computer interface, highlighting the shift from text-only evaluation to pixel-level motor control~\cite{computerUsingAgent2024}.

\paragraph{Exploration and mobile-control benchmarks.}
\noindent\textbf{MobileWorld:} Restores headroom for mobile evaluation with 201 tasks emphasizing longer horizons (average 27.8 steps) and cross-application interactions~\cite{mobileworld2025}.

\noindent\textbf{AndroidWorld:} Established a reproducible, emulated environment for mobile agents, though it has reached a ``saturation point'' for specialized agents on simpler, single-app tasks~\cite{androidworld2024}.

\noindent\textbf{Mobile-Agent-v2:} Mitigates exploration cost by utilizing improved memory and multi-agent coordination to navigate effectively~\cite{wang2024mobile}.

\section{Failure-Taxonomy Details}
\label{app:failure-taxonomy-details}

This appendix records the mapping from MAST's multi-agent categories to the single-agent computer-use taxonomy.

\begin{table}[h]
\centering
\scriptsize
\setlength{\tabcolsep}{3pt}
\renewcommand{\arraystretch}{1.03}
\begin{tabularx}{\linewidth}{@{}X p{1.55cm} X@{}}
\toprule
\textbf{MAST category / mode} & \textbf{Disposition} & \textbf{CUA audit category} \\
\midrule
Disobey task specification & Kept & T1-SPEC \\
Disobey role specification & Removed & Not applicable to one agent \\
Step repetition & Reinterpreted & T1-LOOP if replanning fails; T3-NOOP if feedback is ignored \\
Loss of conversation history & Reinterpreted & T2-STATE \\
Unaware of termination conditions & Reinterpreted & T3-TERM \\
Reasoning--action mismatch & Reinterpreted & T2-GROUND or T2-TOOL \\
Communication breakdowns (reset, ignored input, withholding) & Reinterpreted & Within-agent state propagation: T2-STATE \\
Other intrinsically inter-agent misalignment & Removed & Not applicable to one agent \\
Premature termination & Kept & T3-TERM \\
Incorrect or incomplete verification & Kept/elevated & T3-VERIF; observable no-change becomes T3-NOOP \\
GUI grounding (not explicit in MAST) & Reinterpreted / elevated & T2-GROUND \\
\bottomrule
\end{tabularx}
\caption{Provenance mapping from MAST~\cite{cemri2025multi} to the CUA audit codebook.}
\label{tab:mast-cua-mapping}
\end{table}

\paragraph{Methodology.}
We adapt MAST to the single-agent setting by removing intrinsically multi-agent categories, reinterpreting communication breakdowns as failures of state propagation across steps, and adding GUI grounding and long-horizon verification as first-class computer-use categories.

\section{Future-Direction Details}
\label{app:future-details}

This appendix expands the future-directions discussion of \S\ref{sec:future}: Agent-as-a-Judge methodology, trajectory and efficiency details, and dynamic-benchmark mechanisms.

\subsection{Agent-as-a-Judge}
Methodology: A strong model grades from the user request, the agent's trajectory, and the final output, scoring both workflow quality and outcome correctness~\cite{agentAsJudge2024,bavaresco2025llms}. Performance \& Cost: Agent-as-a-Judge can reach \mbox{$\sim$90\%} agreement with human experts (vs. \mbox{$\sim$70\%} for simple LLM-as-a-Judge prompts) while cutting evaluation cost by \mbox{$\sim$97\%} (e.g., 86 hours to 2 hours)~\cite{agentAsJudge2024}.

Meta-Evaluation: Judge trust must be earned via audits against human-labeled trajectories; benchmarks like AgentRewardBench stress-test reliability (e.g., verbosity bias, self-preference) before judge scores are treated as evidence~\cite{agentRewardBench2025}.

\subsection{Trajectory and Efficiency}
The field is moving from ``outcome-based'' evaluation (did it pass?) to ``process-based'' evaluation (how did it pass?). The TRACE framework makes the trajectory a first-class artifact and validates the step-by-step logic under a validate-by-reproduce requirement~\cite{selfEvolvingBench2025}.

Efficiency Metrics (AgentDiet): As benchmarks shift to reward ``lean'' agents instead of capable ones, AgentDiet prunes redundant or expired context from trajectories. Empirical studies show this reduces input tokens by up to 59.7\% and costs by up to 35.9\% without sacrificing success rates~\cite{agentdiet2025}. This matters most in token-heavy judge-based evaluations ($T_{\mathrm{in}}+T_{\mathrm{out}}$): pruning lowers the marginal cost of verification and transitions complex judge-based audits from ``research-only'' into budgeted, periodic runs for smaller teams.

\subsection{Dynamic Benchmarks}
To combat benchmark saturation and contamination, evaluation environments must evolve faster than the models they test.

(i) AgentGym \& AgentEvol: These frameworks propose environments that ``grow.'' AgentGym provides a suite of diverse, interactive environments, and its AgentEvol method allows agents to evolve by exploring them rather than imitating static datasets~\cite{agentgym2024}. Combined with task-evolution mechanisms such as TRACE's evolutionary proposer~\cite{selfEvolvingBench2025}, this creates an ``arms race'' between the agent and the benchmark, ensuring the evaluation never hits a ceiling.
(ii) Procedural Generation: Future benchmarks will likely use procedural generation to create infinite variations of tasks (e.g., varying file names, UI layouts, and data values in OSWorld) to prevent overfitting and memorization~\cite{wang2019paired}.
(iii) Controlled stochasticity (seeded randomization): To reconcile realism with reproducibility, benchmarks can inject ``seeded randomization.'' By introducing stochastic elements like network lag or transient UI drift via an explicit random seed, labs can test agents against real-world ``mess'' while maintaining deterministic reproducibility. In this framework, system snapshots provide the foundation, while seeded perturbations provide the realism layer.

\section{Scoping-Review Protocol}
\label{app:scoping-protocol}

We characterize our literature process as a \textit{scoping review}: it aims for representative coverage of agent-evaluation practice, not the exhaustive, protocol-registered coverage of a PRISMA-style systematic review.

\paragraph{Sources.} ACL Anthology (ACL, EMNLP, NAACL); OpenReview and official proceedings for NeurIPS, ICLR, and ICML; arXiv (cs.CL, cs.AI, cs.HC); official benchmark repositories on GitHub; and industry technical reports and documentation.

\paragraph{Query strategy.} Four query families, each run across the sources above and complemented by forward and backward citation chasing from seed papers: (i) (``computer-use agent'' OR ``GUI agent'') AND (benchmark OR evaluation OR leaderboard); (ii) (``web agent'' OR ``desktop agent'' OR ``mobile agent'') AND evaluation; (iii) seed-benchmark citation chasing from WebArena, WorkArena, OSWorld, and ScreenSpot-Pro; (iv) (contamination OR ``evaluator bias'' OR ``reward hacking'') AND (``computer use'' OR ``GUI agent'').

\paragraph{Inclusion criteria.} An artifact is included if it (i) evaluates LLM-based agents that perceive and act through graphical web, desktop, or mobile interfaces, or audits such an evaluation; (ii) was released between 2022 and mid-2026; (iii) is available in English; and (iv) provides enough methodological detail to classify its evaluator type.

\paragraph{Exclusion criteria.} We exclude (i) pure text generation, static QA, API-only tool use, and code- or command-line-only benchmarks with no GUI action loop; (ii) multi-agent-only coordination frameworks, which fall outside our single-agent scope; (iii) position papers without evaluable methodology; and (iv) duplicate versions of the same work (arXiv and venue versions counted once, citing the venue version where available).

\paragraph{Corpus.} Screening was performed iteratively alongside drafting rather than in a single logged pass, so we do not report PRISMA-style identification and exclusion counts; per-stage tallies were not recorded, which is a limitation of the protocol (and an instance of the logging discipline that \S\ref{sec:guidelines} asks of benchmarks). The verifiable outcome of screening is the cited corpus itself, whose benchmark-bearing subset is coded in Table~\ref{tab:cua-evaluation-landscape}. Borderline candidates were excluded chiefly under criteria (i) (no action loop) and (iii) (no evaluable methodology).

\paragraph{Coding.} Included artifacts were coded along four dimensions: domain, evaluator type (string/state matching, programmatic state checks, LLM or agent judge, human), failure modes discussed, and mitigations proposed. Table~\ref{tab:cua-evaluation-landscape} preserves the screened GUI-CUA landscape, while the empirical audit centers the benchmarks with released trajectories.

\end{document}